\documentclass[conference]{IEEEtran}
\usepackage{times}

\usepackage[numbers]{natbib}
\usepackage{multicol}
\usepackage[bookmarks=true]{hyperref}
\usepackage{multirow}

\usepackage{hyperref}
\usepackage{url}
\usepackage{float}

\usepackage{comment}
\usepackage{mathtools}
\usepackage{amsfonts}  %
\usepackage{cleveref}
\usepackage{graphics}
\usepackage[pdftex]{graphicx}
\usepackage{wrapfig}
\usepackage[font=footnotesize]{caption}
\usepackage{color}
\usepackage{dsfont}
\usepackage[]{mdframed}
\usepackage{algorithm}
\usepackage{algorithmic}
\usepackage{xparse}
\usepackage{amsmath}
\usepackage{bm}
\usepackage{mathtools}
\usepackage{amssymb}
\usepackage{amsthm}
\usepackage{amssymb}

\usepackage{booktabs}
\usepackage{epigraph}
\usepackage{listings}
\usepackage{xcolor} %
\usepackage{subcaption}
\usepackage{xspace}
\usepackage{capt-of} %
\usepackage{cuted}   %

\lstdefinestyle{pythonstyle}{
    language=Python,
    basicstyle=\small\ttfamily,
    keywordstyle=\color{blue},
    stringstyle=\color{red},
    commentstyle=\color{green!50!black},
    morecomment=[l]{\#},
    showstringspaces=false,
    numbers=left,
    numberstyle=\tiny\color{gray},
    frame=single,
    breaklines=true,
    tabsize=4
}

\usepackage{xcolor} %
\usepackage{hyperref} %
\hypersetup{
    colorlinks=true,
}

\usepackage{xargs} %
\usepackage[colorinlistoftodos,prependcaption,textsize=tiny]{todonotes}
\newcommandx{\wrn}[2][1=]{\todo[linecolor=red,backgroundcolor=red!25,bordercolor=red,#1]{#2}}
\newcommandx{\cmt}[2][1=]{\todo[linecolor=blue,backgroundcolor=blue!25,bordercolor=blue,#1]{#2}}

\newcommand{\method}{$\pi_{0.5}+\text{ego}$\xspace}
\newcommand{\humandata}{embodied human data\xspace}

\def \pifive {$\pi_{0.5}$\xspace}

\def \H1{H1}
\def \G1{G1}

\IEEEoverridecommandlockouts

\begin{document}

\title{
Emergence of Human to Robot Transfer in Vision-Language-Action Models
}

\pdfinfo{
   /Author (Physical Intelligence)
   /Title  (@title)
   /Subject (Robot Foundation Models)
   /Keywords (Robot Foundation Models)
}

\def\cameraready{0}  %

\ifx\cameraready\undefined
    \author{
    Anonymous Submission
    }
\else
\author{
  Simar Kareer$^{1\,2\,*}$ \quad
  Karl Pertsch$^{1}$ \quad
  James Darpinian$^{1}$ \quad
  Judy Hoffman$^{2}$ \\
  Danfei Xu$^{2}$ \quad
  Sergey Levine$^{1}$ \quad
  Chelsea Finn$^{1}$ \quad
  Suraj Nair$^{1}$ \\[4pt]
  $^{1}$Physical Intelligence \quad
  $^{2}$Georgia Institute of Technology
  \\[4pt]
}
    
\fi

\maketitle

\begin{strip}
\centering
\vspace{-4em}
\includegraphics[width=\textwidth]{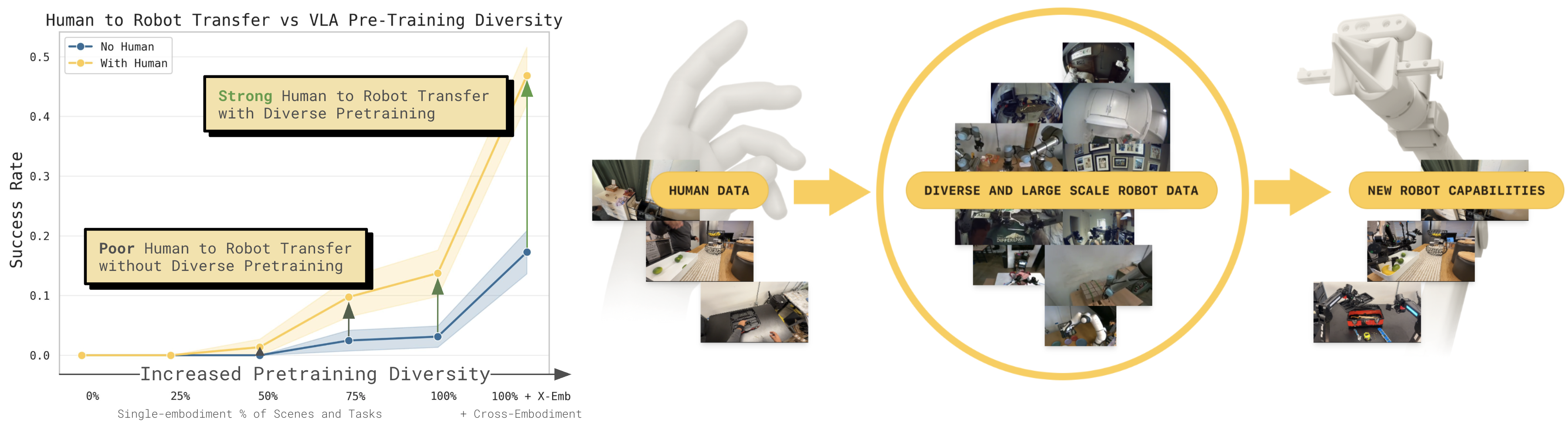}

\captionof{figure}{\textbf{Emergence of human to robot transfer:}
We observe that the transfer from human data to robot policies scales with the size and diversity of VLA pre-training data.
The x-axis represents the diversity of the pre-training robot dataset, and the yellow and blue lines show the finetuning performance with and without human embodiment data. While both increase, the gain from leveraging human data only appears beyond a certain pre-training scale. We evaluate on a suite of four generalization scenarios shown only in the human data.}
\label{fig:teaser}
\end{strip}

\begin{abstract}
Vision-language-action (VLA) models can enable broad open world generalization, but require large and diverse datasets. It is appealing to consider whether some of this data can come from human videos, which cover diverse real-world situations and are easy to obtain. However, it is difficult to train VLAs with human videos alone, and establishing a mapping between humans and robots requires manual engineering and presents a major research challenge.
Drawing inspiration from advances in large language models, where the ability to learn from diverse supervision emerges with scale, we ask whether a similar phenomenon holds for VLAs that incorporate human video data.  We introduce a simple co-training recipe, and find that human-to-robot transfer emerges once the VLA is pre-trained on sufficient scenes, tasks, and embodiments. Our analysis suggests that this emergent capability arises because diverse pretraining produces embodiment-agnostic representations for human and robot data. We validate these findings through a series of experiments probing human to robot skill transfer and find that with sufficiently diverse robot pre-training our method can nearly double the performance on generalization settings seen only in human data.  

\end{abstract}

\IEEEpeerreviewmaketitle

\begin{figure*}[t!]
\centering
\includegraphics[width=\linewidth]{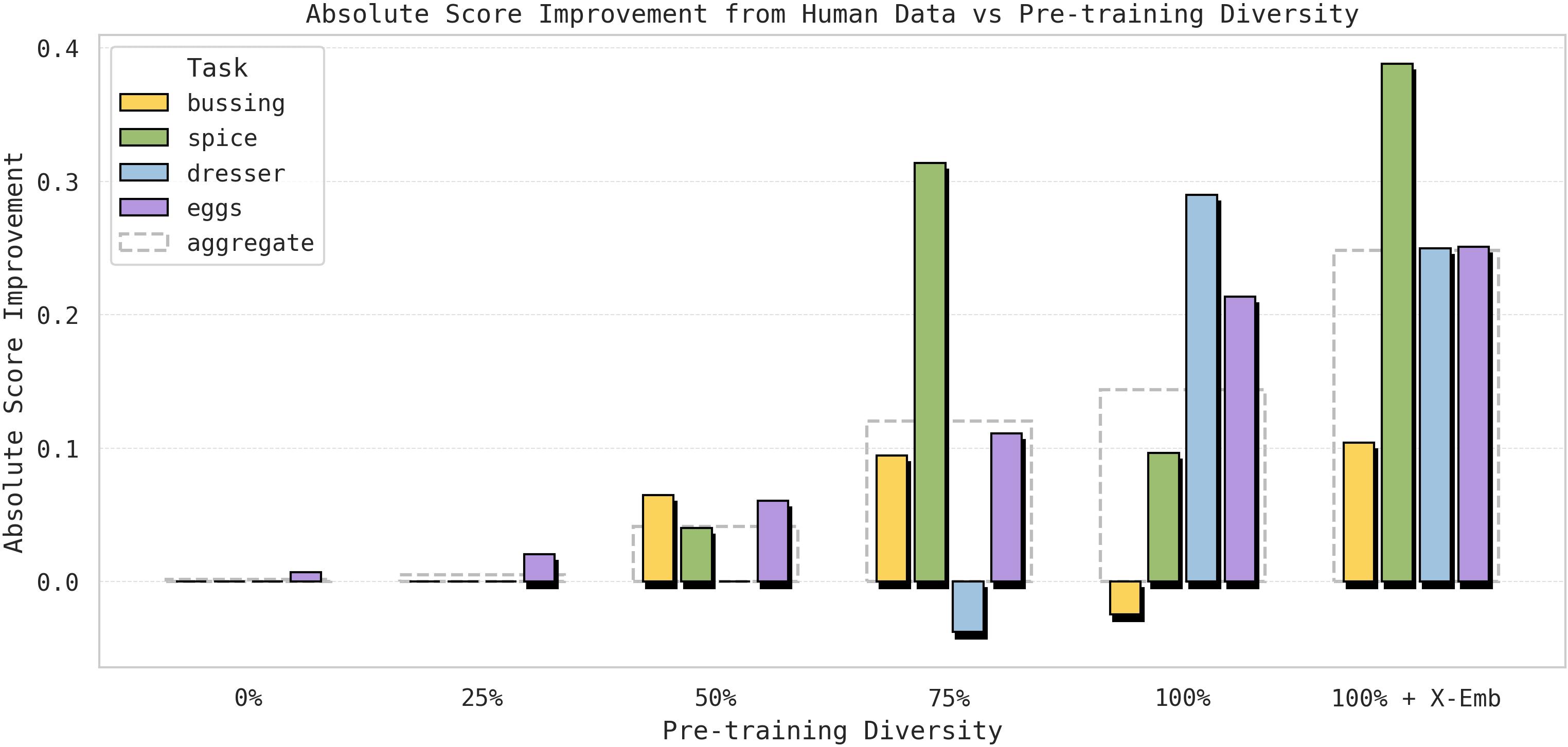}
\caption{\textbf{Per-task improvement from human data:} We plot the \textit{difference} in performance between policies fine-tuned with robot + human data versus robot-only data, isolating the lift from human supervision. Gains are largest when pre-training spans diverse tasks, scenes, and embodiments, suggesting that broad pre-training improves transfer from human videos.}
\label{fig:scalingAll}
\vspace{-10pt}
\end{figure*}

\renewcommand{\thefootnote}{}%
\footnotetext{$^*$Work conducted during internship at Physical Intelligence.}
\renewcommand{\thefootnote}{\arabic{footnote}}%

\section{Introduction}

Human knowledge provides the foundation to instill physical intelligence in robots.  This manifests in many forms, from bootstrapping robot policies with human generated text and images via vision-language models, to mimicking human generated actions via robot teleoperation. While such techniques \textit{indirectly} imbue the model with human experience, the right recipe to directly learn from human experience, for instance by watching a video of someone perform a task, remains an active area of research \cite{chen2021learning, bahl2022human, nair2022r3m, bharadhwaj2024track2actpredictingpointtracks, egomimic, lepert2025phantomtrainingrobotsrobots}. Techniques for leveraging this type of data hold the promise to unlock massive scale human data for generalist robot policies.

Drawing inspiration from language models, recent work has found that the \textit{ability} to leverage certain sources of data is intrinsically tied to model scale \cite{wei2021finetuned, wei2022chain, wei2022emergent}. For instance, while smaller models fail to leverage diverse instruction tuning datasets, larger models become generalists that soak up diverse data and generalize to new tasks.  This leads us to ask: \textbf{does learning skills from human video data, with no explicit alignment, emerge with scale?} To test this hypothesis, we introduce a simple co-training recipe, that treats human videos as an additional embodiment with the same objectives used for robot data. Specifically, we predict low level end effector trajectories using 3D hand tracks and high-level sub-tasks using dense language annotations, mirroring the objectives used during robot pretraining. We then co-finetune on a mix of this human data with the relevant robot data, and evaluate in a setting present only in the human data. 
For example one such setting involves sorting eggs, where the robot data covers placing eggs in cartons, while human data specifies how differently colored eggs should be sorted across multiple cartons.

With this recipe, we uncover our key finding: human to robot transfer is an emergent property of diverse VLA pretraining (\Cref{fig:teaser}). As we expand the diversity of robot data -- across tasks, scenes, and embodiments -- the pretrained VLA becomes increasingly capable of leveraging human videos during post-training. We quantify this effect on four generalization benchmarks that probe different axes of transfer, including unseen apartments, novel object categories, and new task semantics.  Our full recipe leverages human video data to enable capabilities never shown in robot data, for instance it busses unseen objects, tidies an unseen home and performs a task with novel semantic structure. 

These findings might lead one to ask: \textit{why} does diverse pretraining matter so much for transfer? We find that as pretraining diversity increases, the latent representations between human and robot data naturally align.  This suggests that with sufficient data coverage, models begin to form embodiment agnostic representations, despite vast visual and kinematic domain shifts.  In the same way large language models become generalists that can learn from diverse supervision, diverse VLAs become generalists that can learn from diverse embodiments.

Concretely we show that robotic foundation models are able to directly leverage human data when pre-trained on sufficiently diverse data, which we demonstrate quantitatively by performance on generalization tasks, as well as qualitatively by analyzing the structure of the latent embeddings across embodiments.  To test the efficacy of our recipe, we ablate the importance of each training objective, the importance of wrist cameras, and compare the relative value between human and robot data.  We believe this research provides a new perspective on the potential role of \humandata in training state-of-the-art VLAs.  Rather than developing bespoke algorithms to leverage human data, we can think of it in the context of cross-embodiment transfer, where its usefulness is amplified by diverse pretraining.

\begin{figure*}[t!]
\centering
\includegraphics[width=\linewidth]{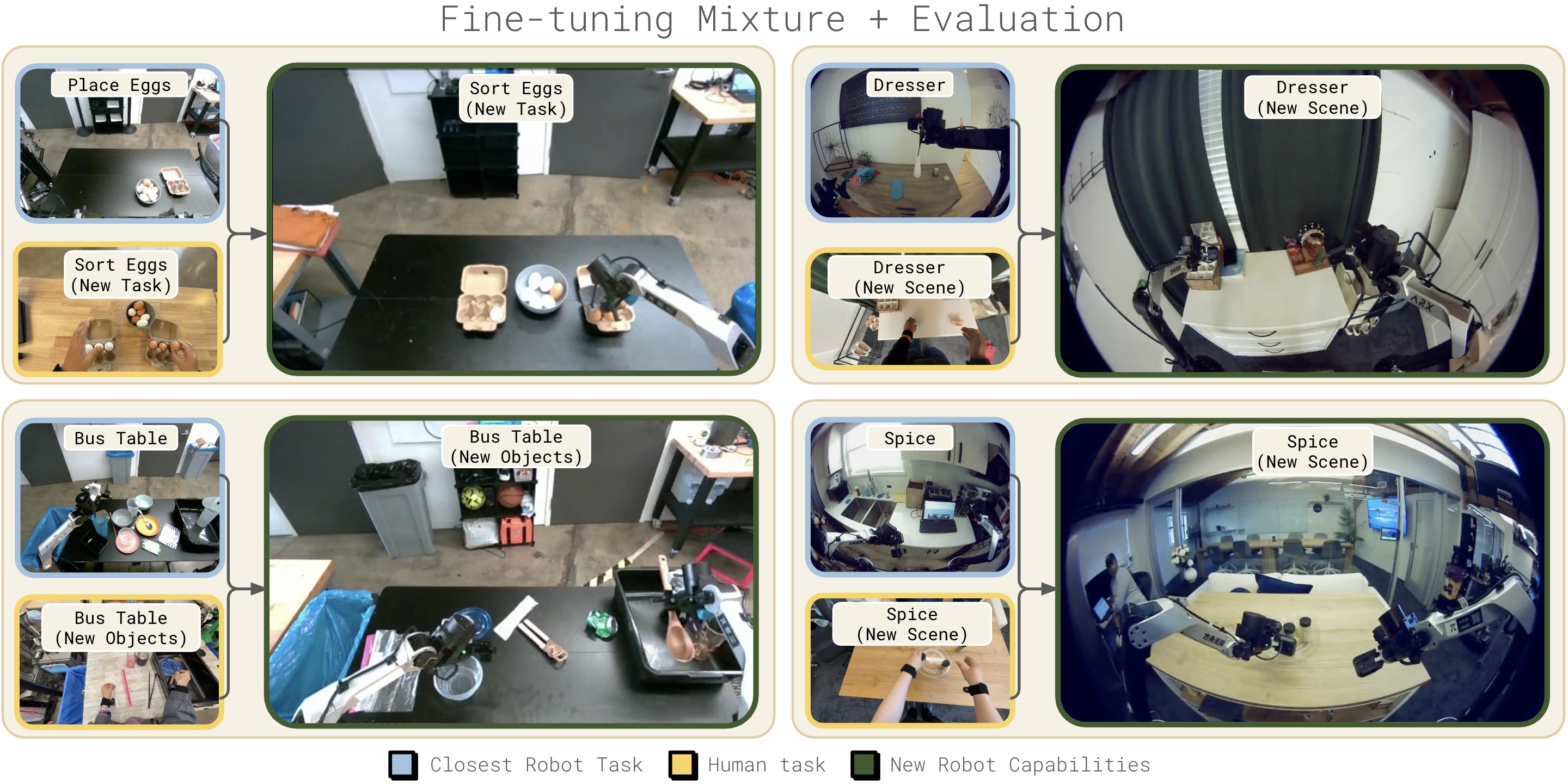}
\caption{\textbf{Training mixture and benchmark.}  Our fine-tuning mix is evenly split between human data for generalization tasks and robot data for the nearest neighbor task.  For each task we evaluate generalization to a new concept introduced only in human data 1) Scene generalization: We have robot data for tidying dressers and spice racks across many airbnbs and human data for an unseen apartment.  2) Object generalization: robot data covers bussing a table filled with trash and dinnerware and human data for a new set of objects.  3) Task generalization: robot data covers placing eggs into cartons, but human data introduces the new concept of sorting eggs by color.}
\label{fig:data}
\vspace{-10pt}
\end{figure*}

\section{Related Works}
\noindent\textbf{Learning from Humans.}
Learning manipulation from human video has received significant attention due to its potential scalability.  Over the years, advances have been made to leverage this data more directly for policy learning. Early works in this field leveraged human video data to train stronger vision encoders, which can improve downstream policy learning~\cite{nair2022r3m, vc2023, ma2022vip}.  Such approaches leverage the visual diversity of large human datasets like Ego4D~\cite{grauman2022ego4d} to train rich visual features, but are unable to directly improve action prediction.  To address this, a number of works developed proxies for actions via intermediate prediction tasks such as keypoint tracking~\cite{bharadhwaj2024track2actpredictingpointtracks, wang2023mimicplaylonghorizonimitationlearning}, latent actions~\cite{ye2024latent}, reward modeling~\cite{chen2021learning}, and affordance prediction~\cite{Bahl_2023_CVPR, bahl2022human}. Alternatively, other approaches use overlayed robots and AR/VR to explicitly align the human and robot actions ~\cite{duan2023ar2d2trainingrobotrobot, park2024dexhubdartinternetscale}. While these works get closer to capturing real human actions, they introduce a manually engineered structure to enable transfer, limiting the generality of tasks that can be captured.

In parallel to this work, advances in AR/VR enable us to extract explicit actions from humans in the form of 3D hand and head tracking~\cite{engel2023projectarianewtool}.  Recent works leverage this advancement to train unified policies on human and robot data with a single objective: future action prediction, whether that be from a human hand or robot end-effector~\cite{egomimic, egobridge, emma, lepert2025phantomtrainingrobotsrobots, liu2025egozerorobotlearningsmart, yang2025egovlalearningvisionlanguageactionmodels, qiu2025humanoidpolicyhuman}.  These works offer a promising path to directly leverage large scale human data, but such methods are generally brittle at a small scale.  As such, they often rely on some form of alignment to work well, whether that be kinematic, visual, or latent.  In our work, we extend this style of work and perform no explicit alignment steps.

\noindent\textbf{Heterogeneous Vision-Language-Action Models.}
Modern VLAs are trained as generalist policies with heterogeneous supervision, combining robot teleoperation, web-scale vision--language data, and language annotations into a single model~\cite{driess2023palm, rt22023arxiv, kim2024openvla, black2024pi_0, wen2024tinyvlafastdataefficientvisionlanguageaction, zhen20243dvla, belkhale2024minivla, szot2024multimodal, pertsch2025fast, geminirobotics2025, wen2025dexvla, bjorck2025gr00t, intelligence2025pi05visionlanguageactionmodelopenworld}. These models bootstrap strong vision--language backbones to get broad semantic understanding from human-generated images and text, and then ground that understanding in robot experience via behavior cloning on large teleoperation datasets~\citep{dasari2019robonet, ebert2021bridge, walke2023bridgedata, open_x_embodiment_rt_x_2023, khazatsky2024droid, bharadhwaj2023roboagent, fang2024rh20t, shafiullah2023bringingrobotshome, contributors2025agibotworld, lee2025molmoact, jiang2025galaxea}. While supervision from web images, videos, and language further improves open-world generalization, this data lacks explicit actions and is visually out-of-distribution for a robot's egocentric observations.

A common theme in recent VLAs is cross-embodiment training~\citep{open_x_embodiment_rt_x_2023, octo_2023, Doshi24-crossformer, yang2024pushing, zheng2025x}, where a single policy is used to control many different robot embodiments with a unified architecture and action representation. These multi-robot VLAs show that skills can transfer across embodiments, often without bespoke alignment beyond shared observation and action spaces. This suggests that heterogeneous, multi-robot pretraining can produce internal representations that are naturally conducive to transfer across embodiments.

We build on this cross-embodiment hypothesis by treating humans as yet another embodiment within the same heterogeneous VLA training recipe. In contrast to non-embodied human videos one might find on YouTube, we leverage embodied human videos with explicit hand motion and language annotations in the VLA mixture. We find that with sufficient pretraining scale, the resulting VLAs naturally form embodiment-agnostic representations that align human and robot trajectories.

\noindent\textbf{Scaling Alternative Data Collection Strategies.}
While most VLAs are largely driven by robot teleoperation data, there's recent work exploring alternative data collection mechanisms which are more scalable.  A number of works use portable hardware that a user operates with their hands to simulate teleoperation~\citep{song2020grasping, young2021visual}, for instance UMI~\cite{chi2024universal} is a hand-held parallel jaw gripper that tracks its own movement to use as demonstration data.  A number of works expanded on this design to capture data for dexterous hands as well, both via exoskeletons and portable motion capture~\cite{xu2025dexumiusinghumanhand, tao2025dexwilddexteroushumaninteractions, wang2024dexcapscalableportablemocap}.  While these devices are exciting options to increase data scalability, they ultimately encumber the operator, and it is difficult to perform work naturally with these devices

Capturing \humandata offers a promising way to address these limitations, using cameras and computer vision to record 3D hand motion with minimal disruption. This approach enables us to observe human behavior without encumbrance. In this study, we therefore focus on methods for leveraging \humandata.

\section{Preliminaries}

We consider the setting of training generalist policies with vision language action models.  VLAs inherit the architecture and pretrained weights of a vision-language model, but are trained to produce continuous robot control.  VLAs are typically trained via behavior cloning on a dataset of demonstrations $D=(o_t, l_t, a_{t:t+H})$ to produce a policy that maps observation and language command to a trajectory of future actions $\pi_\theta(a_{t:t+H}\ |\ o_t, l_t)$. Actions can either be represented as discrete action tokens~\citep{rt22023arxiv, lee2024behavior, pertsch2025fast}, trainable via standard next-token prediction, or as continuous values, often trained via flow-matching objectives~\citep{black2024pi_0}. In this work, we follow \citet{driess2025knowledge} and train VLA models using \emph{both} action representations: we train our model to predict discretized FAST~\citep{pertsch2025fast} action tokens, and introduce a small action expert network that decodes continuous actions via a flow matching objective. For more details on model architecture and training objective, see~\cite{intelligence2025pi05visionlanguageactionmodelopenworld}.

Recent VLAs like $\pi_{0.5}$ showed improved generalization from \emph{co-training} VLAs with additional objectives like subtask prediction, object detection, and VQA.  For subtask prediction, the policy predicts a subtask string given visual observations and a high level language command $p(l^{subtask}_t\ |\ o_t, l_t)$.  This language is fed back into the model to condition action generation $\pi_\theta(a_{t:t+H}\ |\ o_t, l^{subtask})$, similar to chain of thought.  Subtask labels are obtained by densely annotating demonstration data with language descriptions of short, atomic action sequences.  In this work, we train on human data with two objectives: flow based prediction for continuous actions and language based subtask prediction.

\begin{figure}[t!]
\centering
\includegraphics[width=\linewidth]{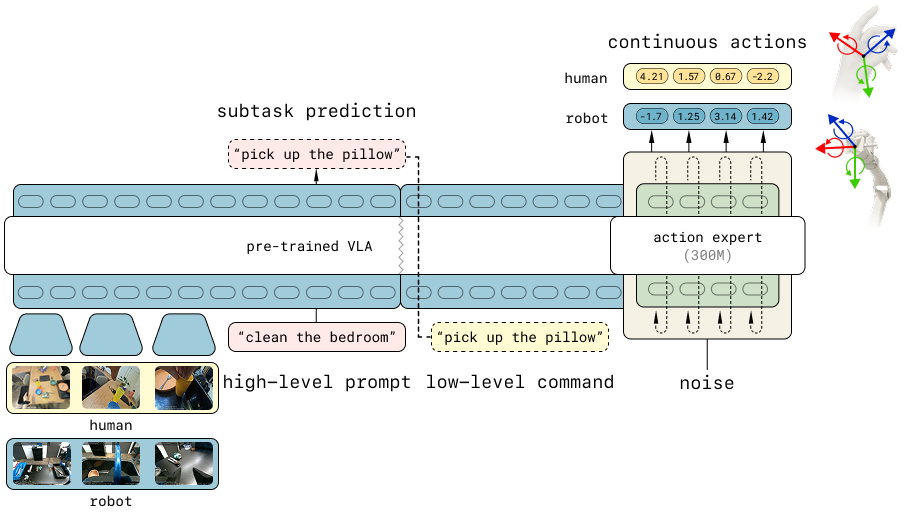}
\caption{\textbf{Model Architecture.} We use the \pifive model.  We finetune with a combination of high level sub-task prediction and low level action prediction on both human and robot data. The low level action prediction leverages relative end-effector actions aligned across human and robot.}
\label{fig:arch}
\vspace{-10pt}
\end{figure}

\begin{figure*}[t!]
\centering
\includegraphics[width=\linewidth]{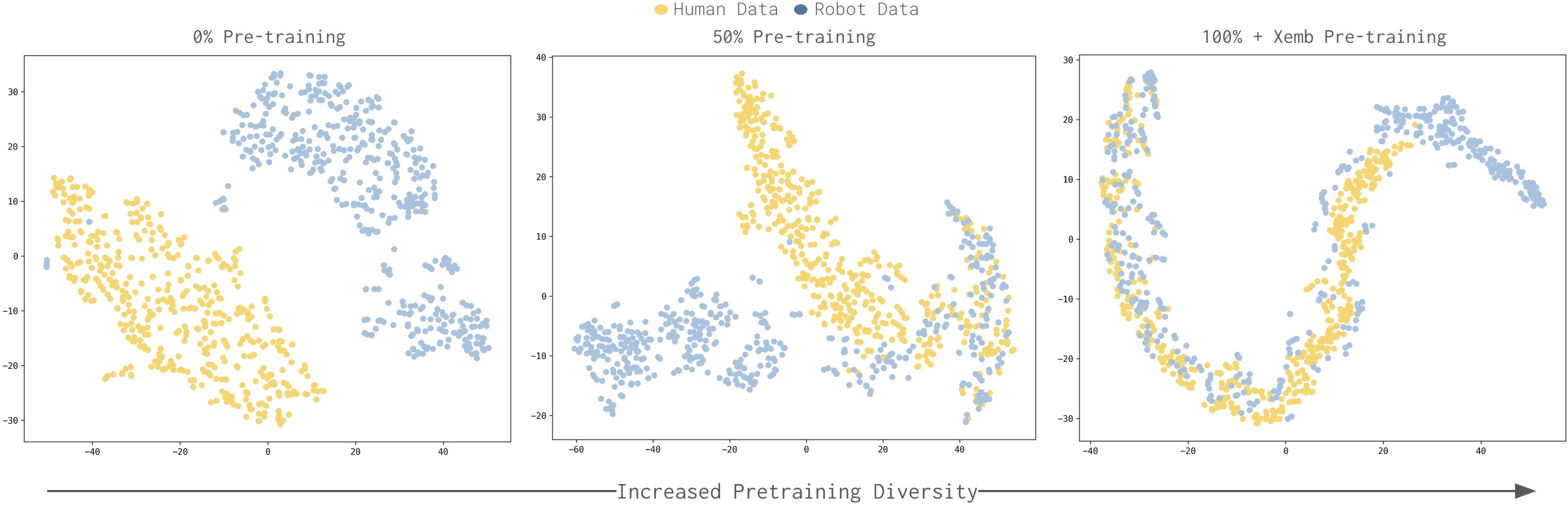}
\caption{\textbf{VLA representation of human and robot data.} We plot the latent embeddings of our VLA by performing a TSNE analysis on mean-pooled tokens from the final layer of the VLM backbone. With no pre-training, it is clear that the model has disjoint representations between human and robot data.  But as pretraining becomes more diverse, latent overlap increases, which correlates with performance on our generalization tasks.  }
\label{fig:tsne}
\vspace{-10pt}
\end{figure*}

\section{Fine-Tuning with Emergent Human-Robot Alignment}
Our fine-tuning recipe aims to leverage embodied human data identically to other robots in our mixture with no explicit alignment.  This approach is maximally general, and leans on the capabilities of large models to ingest relevant information from diverse sources, rather than human designed heuristics for aligning domains. We first collect, process, and annotate the human video data, and then use it in combination with robot data to fine-tune the pre-trained model, which we base on the $\pi_{0.5}$ model shown in \Cref{fig:arch}. The fine-tuning objective treats human and robot data in exactly the same way, without any explicit transfer learning method or loss.

\subsection{Human Data Collection Pipeline}
\noindent\textbf{Data collection device.}
\begin{figure}[t!]
\centering
\includegraphics[width=\linewidth]{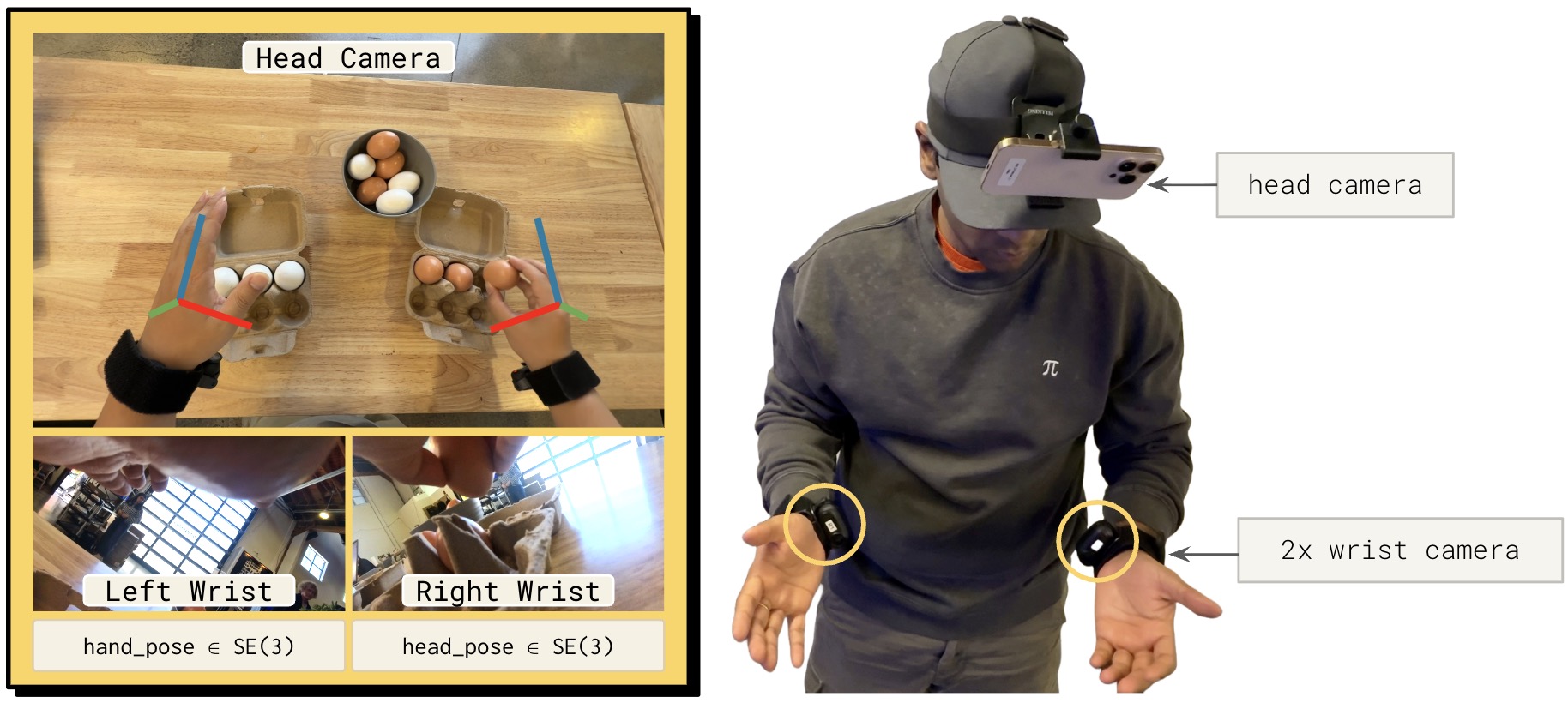}
\caption{\textbf{Data Collection Apparatus.} We collect human data with three views: head mounted camera, left wrist camera, and right wrist camera.}
\label{fig:apparatus}
\vspace{-10pt}
\end{figure}

We design our data collection setup to enable capture of a wide range of human interactions, while being minimally intrusive, and thus scalable. We equip human data collectors with a head-worn, high resolution camera. Since recent robotics research has demonstrated the benefit of wrist-mounted cameras for policy learning, which provide a more detailed view of the end-effector's interactions with manipulated objects, we also experiment with equipping data collectors with wrist-mounted cameras that provide two additional, time-synchronized camera streams. We ablate the effect of these additional cameras in \Cref{sec:experiments}.

\noindent\textbf{Data collection protocol.}
Our intent is to collect human data in the style of episodic robot teleoperation data, which allows us to isolate the transfer problem to only the visual and kinematic differences between human and robot.  As such, operators are instructed to collect repeated demonstrations for each of our tasks while wearing the data collection device.  Further, operators are asked to keep their hands in the field of view of the camera to improve tracking quality.  
We collected 3 hours of data for bussing, 3 hours for spice, 3 hours for dresser and 5 hours for sort eggs.

\noindent\textbf{Data processing \& annotation.}
Given a raw video capture of human interactions, we use visual SLAM to reconstruct the 6D movement $e_t\in\mathbb{R}^6$ of the head-mounted camera relative to a constant world-frame. We also reconstruct the position of 17 3D keypoints $h^{e_t}_t\in\mathbb{R}^{3\times17}$ for both hands in the head camera frame. Finally, analogously to the robot teleoperation data in our training mixture, we annotate the human video data with text-based subtasks, describing the actions of each arm.

\noindent\textbf{Action space.}
We aim to roughly align the action representations for human and robot data in our training mixture to facilitate transfer. For robot teleoperation data, there are multiple choices of action representations.  Two common options are to represent actions as trajectories of robot \emph{joint positions} or \emph{end-effector poses}.  We will consider end-effector based actions, since approximating joint positions for humans is difficult.  Specifically, these end-effector actions are represented as a length  $H$ action chunk $[a_0, a_1, ..., a_H]$, where each $a_i$ represents the 6-DoF pose relative to the current observed state 6-DoF pose $s_0$. 
The total action space for robot data is the concatenation of the 6 DoF left arm end-effector trajectory + gripper, 6 DoF right arm end-effector trajectory + gripper, and 2 dimensional base actions, yielding a total action chunk of $a\in\mathbb{R}^{H\times 16}$.  %
To compute corresponding actions in the human video, we define an ``end-effector'' pose spanning the 3D keypoints of each hand's palm, middle, and ring finger (\Cref{fig:apparatus}),relative to the head frame $e_t$. We then compute end-effector actions as relative transformations from the current 6-DoF state similar to how we do for the robot end effector. 
Similarly, we approximate relative robot base actions by projecting the human video base camera poses into the frame of the chunk's first timestep base camera pose.
We do not explicitly approximate ``gripper actions'' for the human video, since it is challenging to estimate the openness of a human hand during object interactions, and instead rely on learning gripper actions from robot data only. As a result, our human actions have $2\times6 + 6= 18$~dimensions.

\noindent\textbf{Training objectives.}
Our best recipes to perform difficult long horizon tasks leverage both high level subtask prediction and low level action prediction.  We construct both of these prediction tasks on our human data.  For low level action prediction, we supervise action chunk prediction both via next token prediction on discrete FAST tokens as well as flow matching loss on the continuous actions $\pi_\theta(a|o_t, l^\text{subtask}_{t})$.  For high level subtask prediction we train on next token prediction on subtask language tokens $\pi_\theta(l^\text{subtask}_{t}\ |\ o_t, l_t)$.

\noindent\textbf{Training mixture.}
At fine-tuning time, it is important to create a training mixture that both retains the model's original capabilities, while introducing new concepts from human data to improve generalization.  Our mixture reflects this with a simple recipe: we \emph{co-train} human data for generalization tasks at 50-50 proportion to the nearest neighbor robot task.
We use this mixture to finetune $\pi_{0.5}$, a strong VLA exhibiting zero shot generalization, and further improve its capabilities. As shorthand, we refer to the combined model that integrates egocentric data into $\pi_{0.5}$ as \method.

\section{Experimental Findings}
\label{sec:experiments}
To test whether \method can generalize to new concepts from egocentric human data, we construct a suite of ``generalization'' scenarios that have limited coverage in robot data, but present in human data.  These scenarios span generalizing to new scenes, objects and tasks.  We begin our study by understanding whether our recipe can enable transfer to these new settings.  Then, we validate our core hypothesis, which is that this transfer is an emergent property of diverse VLA pretraining.  Finally, we compare human embodiment data to other robot embodiments, study whether transfer occurs from high level subtask or low level action prediction, and ablate the impact of our human-worn wrist cameras.

\subsection{Human to robot transfer benchmark}
Our benchmark aims to test human to robot transfer across various axis of generalization: scene, object and task (\cref{fig:data}).  For each axis, we consider a setup where our robot teleoperation data lacks coverage, and we collect targeted human data to expand this coverage.  In each setting we co-train using \method and evaluate on the new concept introduced in human data.

\noindent\textbf{Scene transfer: }We identified two tasks where we have robot data coverage in a fixed number of homes, but \pifive trained only on robot data fails to generalize to an \emph{unseen} home: \texttt{Spice} and \texttt{Dresser} in which the robot must tidy a spice rack and the top of a dresser respectively.  We collect human data in the unseen target kitchen, then benchmark \method on this new scene. For both of these shorter horizon tasks the score is a binary success rate. 

\noindent\textbf{Object transfer: }Our robot data has coverage of bussing a messy table covered with trash and dinnerware.  We then collect human data which introduces new objects like kitchen tools, then benchmark \method on these new objects.  For this longer-horizon task the score measures the number of correctly placed objects.

\noindent\textbf{Task transfer: }Our robot data has coverage for picking eggs and packing them into a carton.  We collect human data that \emph{sorts} eggs into two cartons based on color and benchmark \method on this new task. For this longer-horizon task the score measures the number of correctly placed eggs.

For more details about task setup and scoring, see \Cref{sec:transfer_benchmark_details}.

\subsection{The \method recipe enables generalization to unseen scenes, objects, and tasks}

\begin{figure}[t]
\centering
\includegraphics[width=\linewidth]{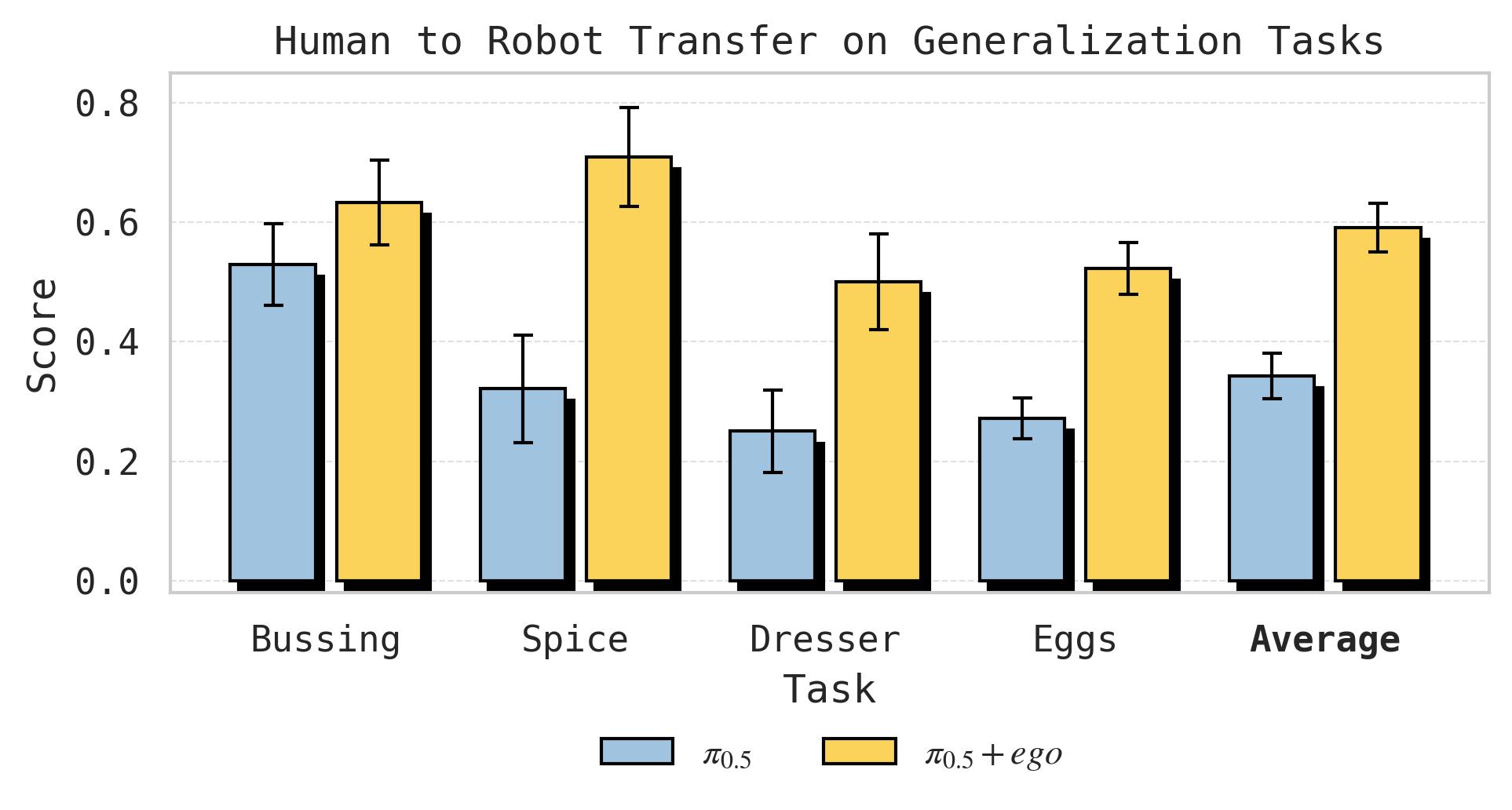}
\caption{\textbf{Human to robot transfer finetuning $\pi_{0.5}$.} We evaluate performance across a suite of static and mobile tasks, each testing either scene, object, or task level generalization present only in the human data. We see clear human to robot transfer resulting in nearly double the score on the target tasks.}
\label{fig:taskPerf}
\vspace{-10pt}
\end{figure}

We report transfer results on our suite of benchmark tasks in \Cref{fig:taskPerf}. \textbf{Across all three generalization axes we find that targeted human data collection and co-training can significantly improve policy generalization.} Concretely, for both scene and object generalization, we see substantially higher task score after co-training: \texttt{Spice}: 32\% $\rightarrow$ 71\%; \texttt{dresser}: 25\% $\rightarrow$ 50\%; \texttt{bussing}: 53$\rightarrow$63\%. Notably, we also see strong \emph{task} transfer from human video in the egg sorting task: while a policy trained on robot data only has the basic manipulation skills to pick and place eggs, it has no concept of \emph{sorting} and simply places eggs into cartons randomly (57\% sorting accuracy). In contrast, once co-trained with human egg sorting videos, the robot policy is able to sort eggs with 78\% accuracy, and on average placed 4~more eggs correctly than \pifive.

\subsection{Human to robot transfer emerges as a function of diverse VLA pretraining across scenes, tasks and embodiments}
We've established that \method can leverage \humandata to expand its capabilities, which leads us to the central question of this work: what enables this transfer?  We hypothesize that policy \emph{pretraining} with a diverse data mix of many scenes, tasks and embodiments is the key enabler for effective human to robot transfer.  Intuitively, VLAs with strong pretraining may learn abstractions that span embodiments—organizing their representations to capture shared structure across domains, and thus facilitating transfer. We test this hypothesis in two parts. First, we establish that human to robot transfer on our generalization benchmark \emph{increases} as a function of pretraining diversity.  Then, we analyze our model's learned representations as pretraining diversity increases.

\noindent \textbf{Strong human to robot transfer emerges with diverse pretraining.}  To evaluate the impact of VLA pretraining on human to robot transfer, we repeat our transfer benchmark experiments but with the following, increasingly diverse, pretrained initializations:
\begin{itemize}
    \item \texttt{0\%}: base VLM initialization only
    \item \texttt{25\%, 50\%, 75\%, 100\%}: VLA pre-trained on increasingly diverse robot data, corresponding to fractions of the full diversity of $[$scene-task$]$ combinations in our data, constrained to the target robot embodiments: ARX and mobile ARX
    \item \texttt{100\% + X-emb}: the \pifive full VLA pretraining mixture~\citet{intelligence2025pi05visionlanguageactionmodelopenworld}, which additionally contains data across numerous non-target robot embodiments.
\end{itemize}

With each of these pretrained initializations we train two models: one with only robot teleoperation data from the most similar tasks in our dataset, and one which additionally includes human embodiment data for these tasks.  This allows us to measure the impact of diverse pretraining on human to robot transfer.

We report results in \Cref{fig:scalingAll}. Concretely, we report the \textit{difference} in score between the model using human data and not using human data at different levels of pre-trained model scale. 
This difference represents the magnitude of the human to robot \textit{transfer} as a function of pre-training diversity. We find that this transfer significantly increases as a function of pre-training diversity. While with no or little pre-training, VLAs cannot benefit from human data co-training (0\%, 25\%), VLAs pre-trained on diverse data see significant gains from human data co-training (75\%, 100\%). Transfer is further improved by pre-training on a diverse cross-embodiment data mix, that includes data from diverse, non-target robot embodiments.

\begin{figure}[h!]
\centering
\includegraphics[width=0.8\linewidth]{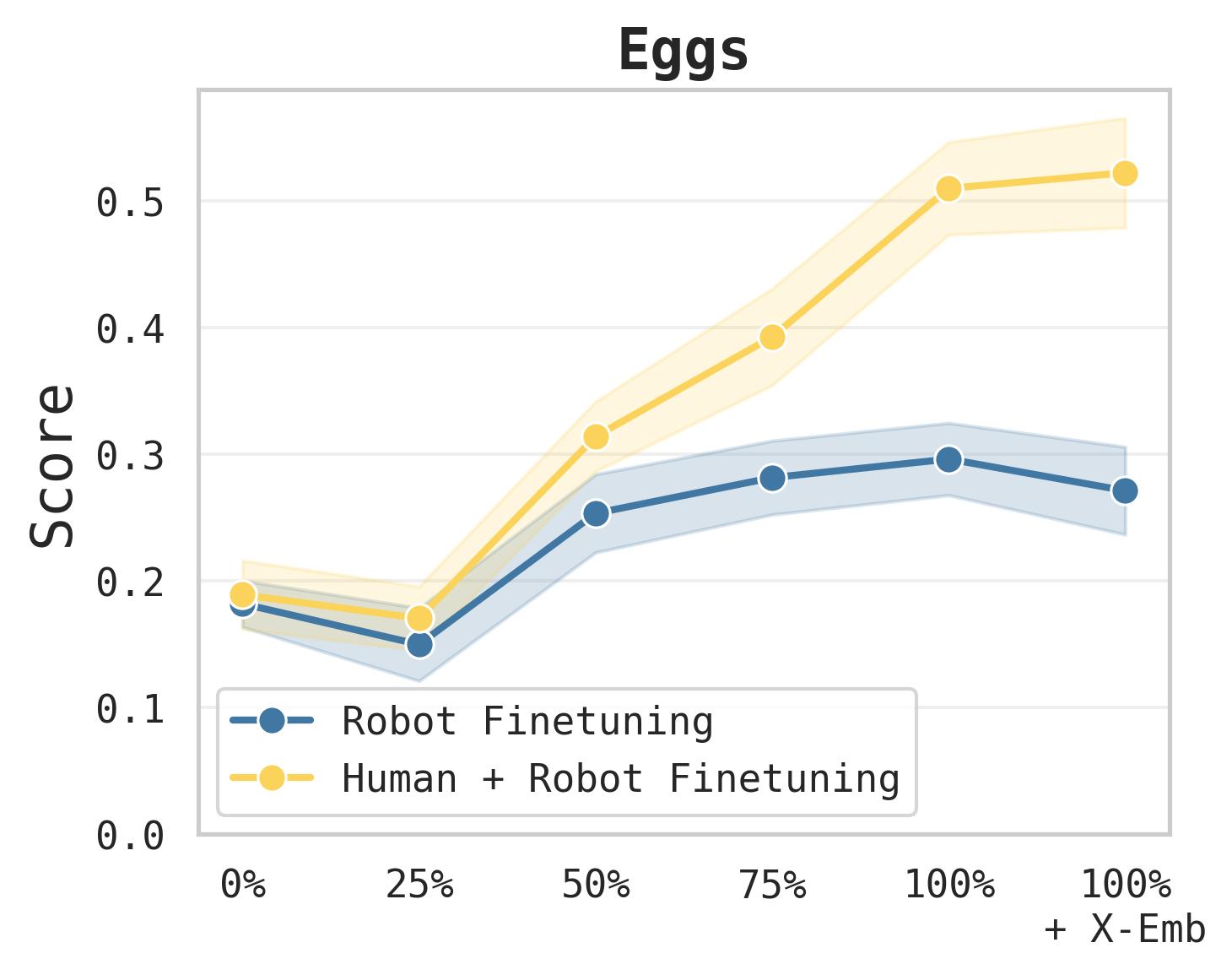}
\caption{\textbf{Task generalization performance scaling:} Performance of Robot Finetuning on \texttt{Sort Eggs} plateaus, even as the pretraining diversity improves. In contrast, Human+Robot Finetuning performance scales sharply with pretraining, suggesting that broader pretraining enables more effective \textit{transfer} from human data.}
\label{fig:scalingEggs}
\vspace{-10pt}
\end{figure}

We can analyze the scaling trends for each task individually.  
For instance in \texttt{Sort Eggs} we see that increased pretraining diversity alone does not enable a robot-only policy to perform \texttt{Sort Eggs}, a task never seen in our robot teleoperation data~(\Cref{fig:scalingEggs}).  However, increased pretraining diversity enables us to \emph{transfer} significantly more knowledge from human data, which does cover this new task.  Similarly, for \texttt{Dresser}, up until our 50\% pretrained checkpoint there are no gains from human video co-training, and potentially even negative transfer (\Cref{fig:scalingAllLines}).  But between \texttt{75\%}$\rightarrow$\texttt{100\% + X-emb} we see consistent stacked gains on top of the robot-only baseline, even as the pretrained checkpoint gets stronger.  More broadly, these results suggest that human to robot transfer will continue to improve as the diversity of our pretrained model grows.  This follows our intuition, since we expect that diversity of scenes, tasks and embodiments ought to improve the model's ability to form embodiment agnostic abstractions.

\noindent\textbf{Embodiment agnostic representations emerge with pre-training scale.}
We hypothesize that diverse pre-training helps produce embodiment agnostic representations which in turn improve human to robot transfer.  To probe this, we perform a TSNE~\citep{tsne} analysis on the output embeddings of the VLA from both human and robot data after co-training (\Cref{fig:tsne}).  With poor pre-training, the model has \emph{disjoint} representations across embodiments, suggesting that the model separately fits these distributions.  Then, as pretraining diversity increases, the representations converge, suggesting the model builds a unified representation for both embodiments. 
See Appendix~\ref{sec:tsne_details} for more details.

Prior works that operate on less data observe that co-training improves performance, but the representations of human and robot are disjoint, and propose methods to explicitly improve representational alignment~\citep{egobridge}.  Our analysis suggests that with sufficiently diverse pretraining, \emph{co-training alone} can produce aligned representations that facilitate transfer.

\subsection{How does \humandata compare to data from other robots?}
\begin{figure}[t]
\centering
\includegraphics[width=\linewidth]{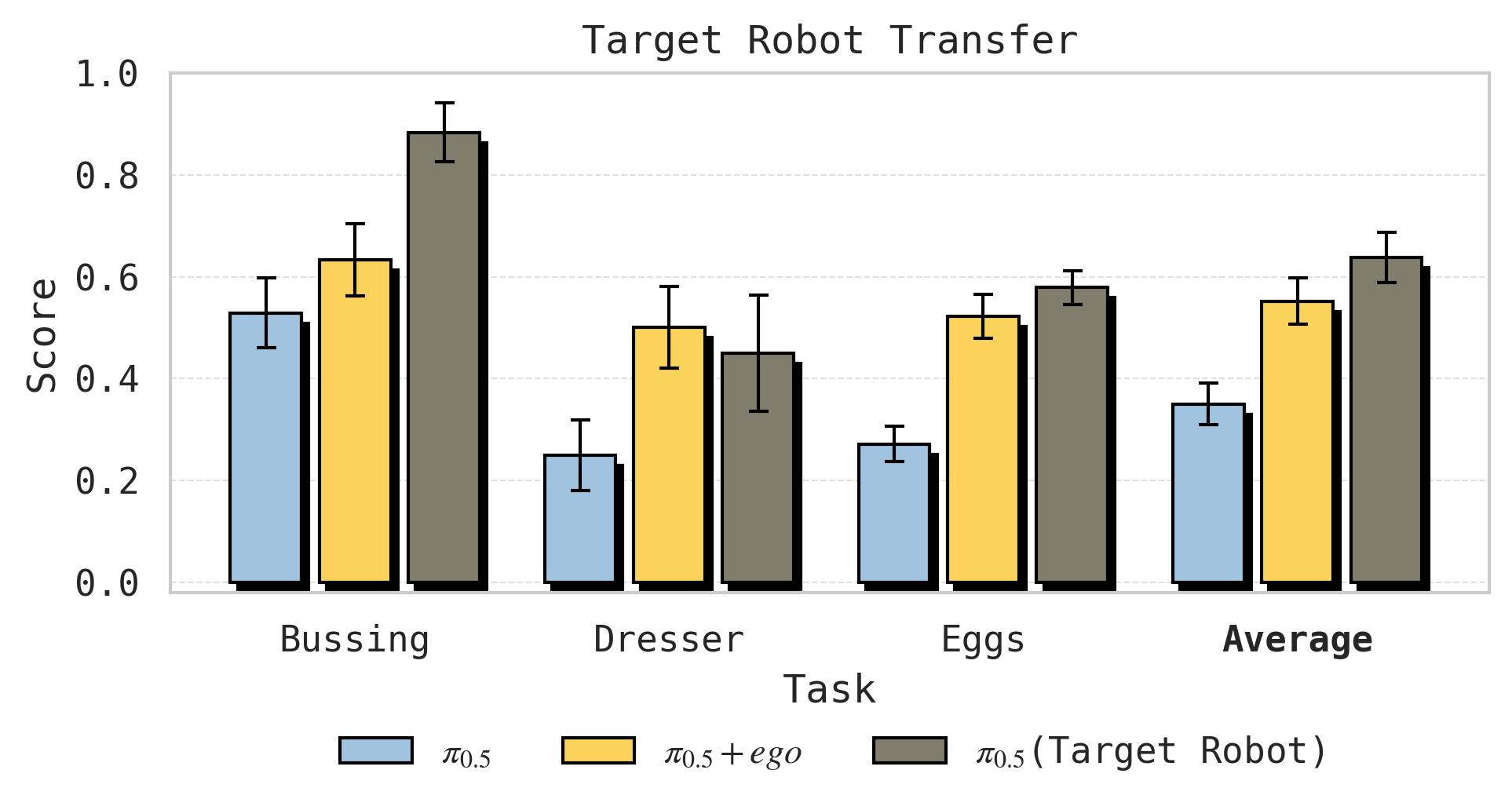}
\caption{\textbf{Human data compared to target robot data:} For \texttt{Eggs} and \texttt{Dresser} we find that fientuning with comparable amounts of human data (yellow) and robot data (grey) resulted in similarly performant models. For \texttt{Bussing} we observe a larger performance gap to target robot data.}
\label{fig:target}
\vspace{-10pt}
\end{figure}

\begin{figure}[h!]
\centering
\includegraphics[width=\linewidth]{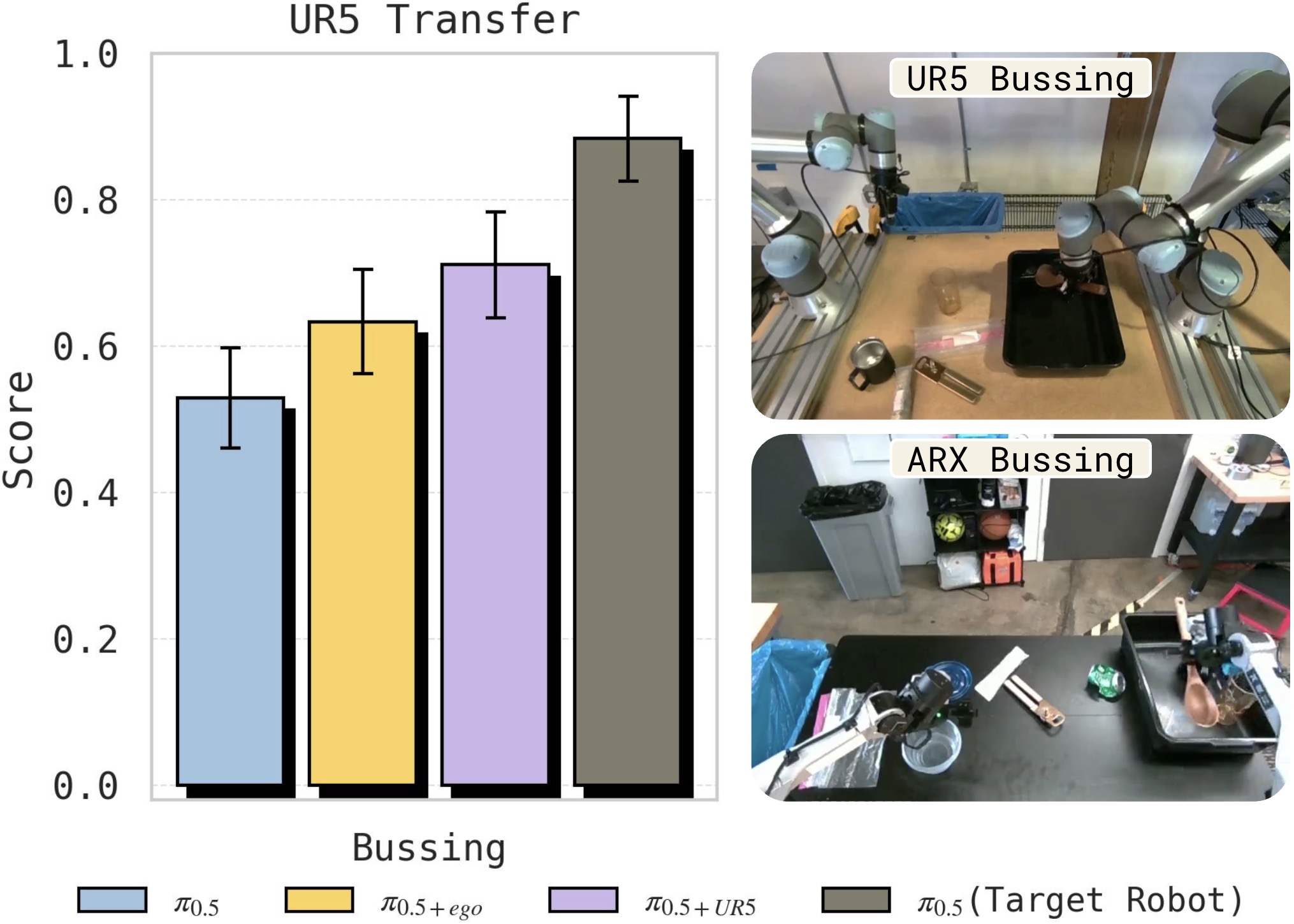}
\caption{\textbf{Human data compared to cross embodiment robot data:} We compare the transfer from human data on the bussing task to that from a robot (albeit a different UR5 robot). We find that both enable lift over the baseline, but neither match data of the target robot, suggesting that human data transfer parallels cross-embodiment robot transfer.}
\label{fig:ur5}
\vspace{-10pt}
\end{figure}

\method frames human to robot transfer as an instantiation of cross embodiment transfer, so it's natural to benchmark human to robot transfer with robot to robot transfer.  This helps us understand whether we can leverage human data as just another robot embodiment in our mix.  We first compare human data to an "upper bound" scenario where we collect target robot data for our benchmark tasks (\Cref{fig:target}).

For two of three tasks (\texttt{Sort Eggs} and \texttt{Dresser}), we find that finetuning with human data were nearly as effective as finetuning with in domain data from the target robot itself.  However, we note that on the \texttt{Bussing} task, target robot data was more effective than human data alone (25\% vs 65\%).

Next we study whether \humandata has roughly the same value as non-target robot data for a new task.
Specifically, for the \texttt{Bussing} task we collected 400 demonstrations (7.45 hours) on a UR5 robot and evaluated transfer to an arx robot.  We see a similar trend in the nature of human to ARX and UR5 to ARX transfer -- both exceed the baseline, but both also don't match data from the target robot embodiment, suggesting that human data transfer and cross-embodiment robot transfer share similar properties.

\subsection{At what level does transfer occur?}
\begin{figure}[t]
\centering
\includegraphics[width=\linewidth]{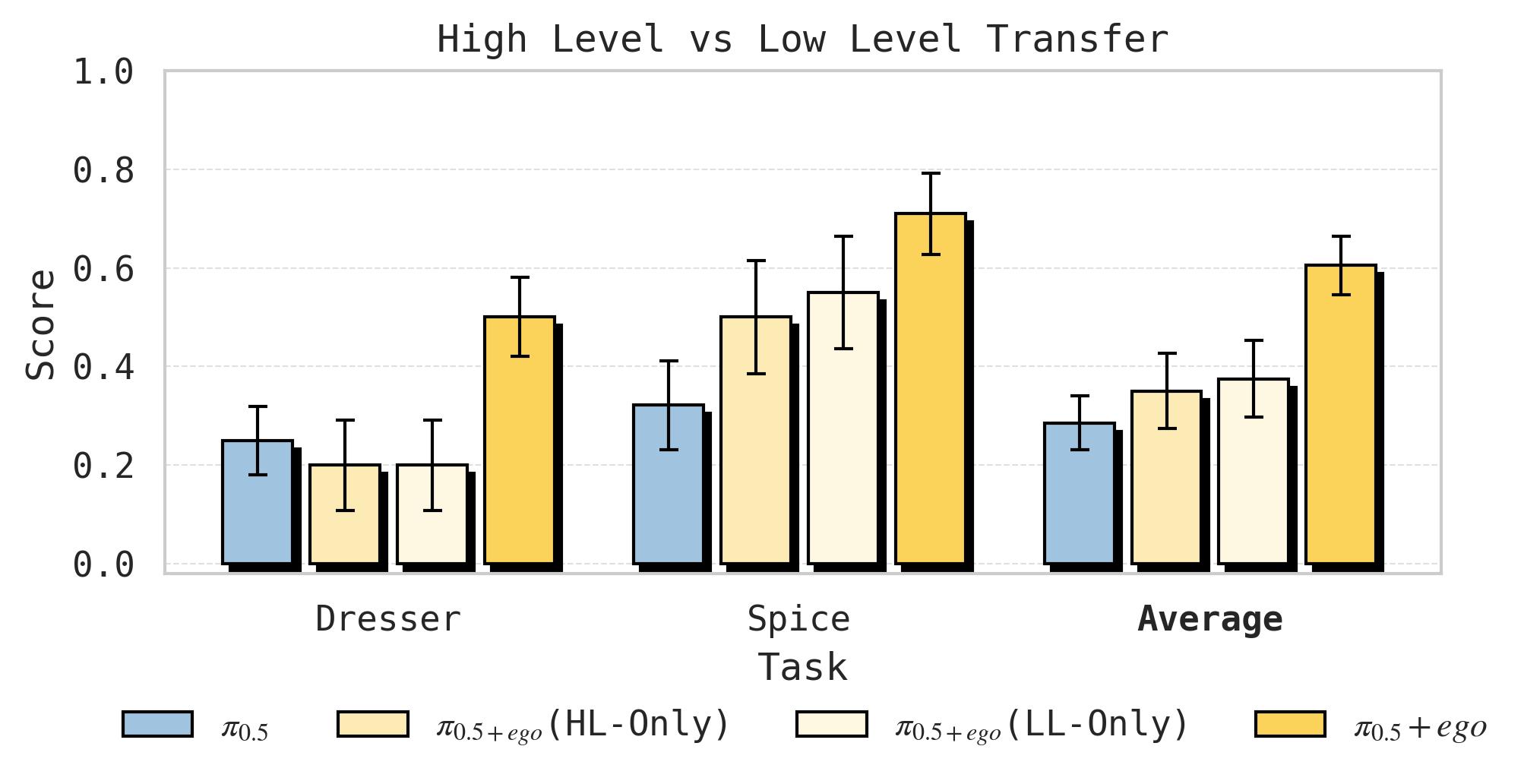}
\caption{\textbf{High level vs low level transfer} For our mobile tasks we find that \humandata enables transfer both through high level sub-task prediction and low level action prediction.}
\label{fig:hl_ll}
\vspace{-10pt}
\end{figure}

A natural question is whether human data can only be used to transfer     ``high level'' semantic concepts, or whether it's also transferring the "low level" action prediction.  For the \texttt{Bussing} and \texttt{Eggs} task, we do not use a high-level policy during evaluation, so the transfer has to come from low level action prediction.  For our mobile tasks \texttt{Spice} and \texttt{Dresser}, we evaluate a joint high level + low level and ablate the impact of each (\Cref{fig:hl_ll}), by testing robot-only HL+LL, robot-only HL + cotrained LL, cotrained HL + robot-only LL and cotrained HL+LL.  We find that leveraging human data for only the HL or LL policy alone is not as effective as co-training both with human data, suggesting that transfer occurs across both levels.  

When we only leverage human data for the HL policy, the low level policies don't follow commands correctly.  For instance in the \texttt{spice} task we observe a failure mode where ``pick up the spice bottle'' is misinterpreted by the low level policy to pick up bottles that are already on the tray.  Or in the \texttt{dresser} task, when the HL says to "put the necklace in the jewelry box", it sometimes puts it in the dresser organizer.

Likewise, when we only leverage human data for LL policy, we get poor HL policy commands.  For instance in \texttt{spice} the high level policy continues to predict ``pick up spice bottle'' long after the bottle has been picked, blocking task progress.  And in \texttt{dresser} the HL policy often predicts incorrect actions, like ``put the hair clip on the top of the dresser'' instead of correctly predicting to put it in the organizer.%

\subsection{How important are the human worn wrist cameras?}%
\begin{figure}[t]
\centering
\includegraphics[width=\linewidth]{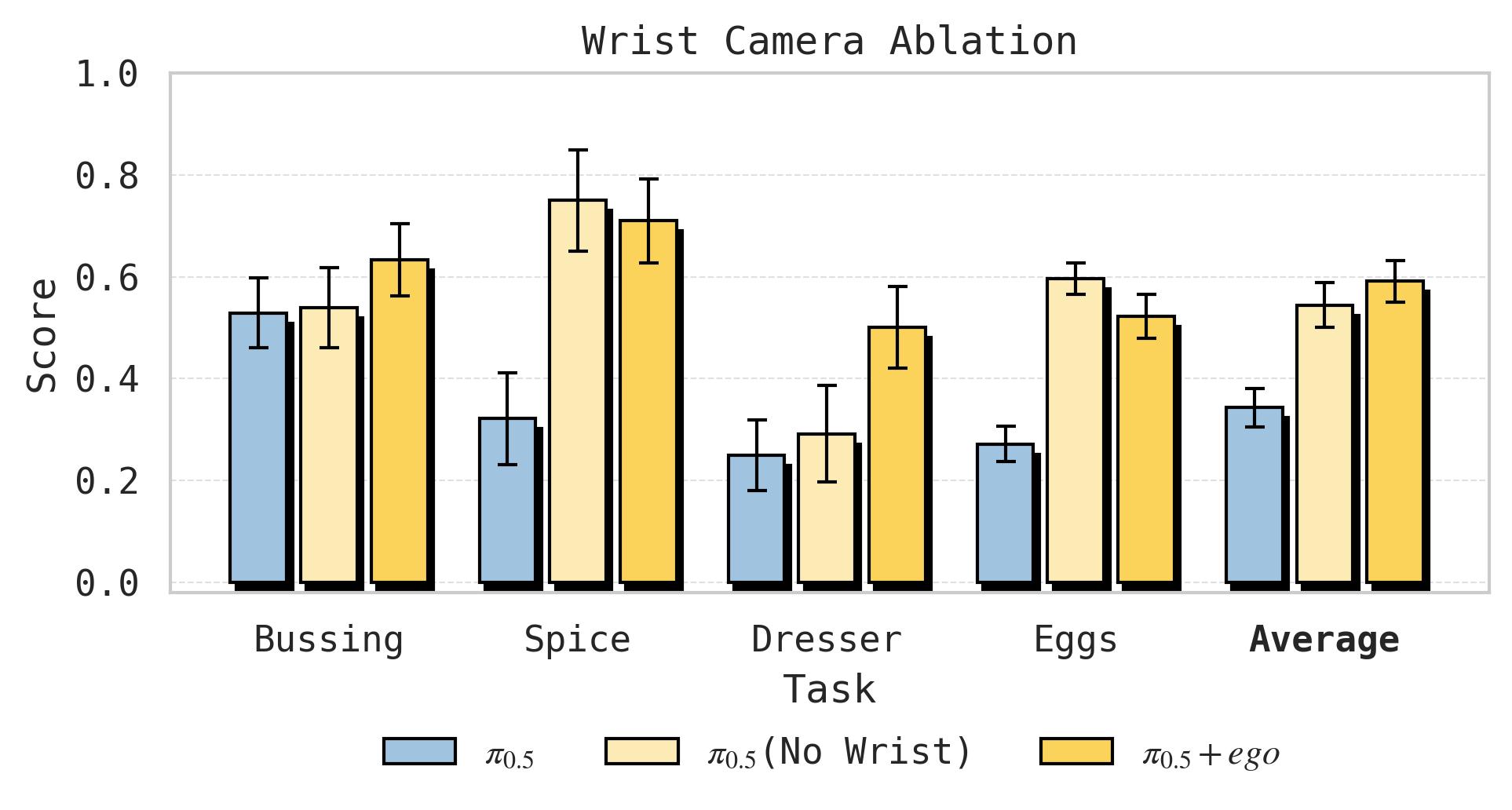}
\caption{\textbf{Human worn wrist cameras:} We see that wrist cameras provide benefit for the  \texttt{Dresser} and  \texttt{Bussing} tasks and had similar performance elsewhere. This matches our intuition that some (but not all) tasks will benefit from the added observability of wrist cameras.}
\label{fig:wrist}
\vspace{-10pt}
\end{figure}

To help mitigate the sensor gap between human and robot, we opted towards collecting human data with small wrist-worn cameras, to mimic the wrist cameras on our robot arms.  We seek to understand their importance, since it has implications on how to sensorize humans for large scale data collection.  We report transfer results with and without wrist camera observations for the human video data in \Cref{fig:wrist}. For some tasks like \texttt{Bussing} and \texttt{Dresser} we see improved transfer from leveraging the human worn wrist cameras, while in other tasks like \texttt{Spice} and \texttt{Eggs}, transfer does not benefit from the additional camera streams.  This matches our expectations, since some tasks rely on wrist cameras for observability more than others.  As a result of these experiments, we expect that collecting embodied human data \emph{with} wrist worn cameras maximally covers the space of potential tasks.

\section{Discussion and Future Work}
We study the emergence of human to robot transfer in our proposed recipe \method. We find that with limited pretraining diversity, VLAs fail to transfer knowledge from human data, but as pretraining diversity grows past a critical threshold, transfer emerges.  While our recipes leverage vast datasets of robot teleoperation data in pretraining, we ultimately only use 10s of hours of human data, and this data is collected in an episodic manner.  We're moving towards a future with vast datasets of embodied human data, which cover the episodic collection like in this work, but also passive data of people performing everyday tasks.  There is additional work to be done to effectively leverage this data during pre-training, but we believe our work lays the groundwork to train VLAs with human data at scale.

Our findings on the emergence of human-to-robot transfer point to a promising future for scaling vision-language-action models. Much like large language models, larger VLAs may not only improve performance but also unlock entirely new capabilities. These capabilities could make it easier to tap into previously hard-to-use data sources and enable more effective transfer across domains—ultimately allowing robotic foundation models to scale \textit{even further}. Using human video may be just one such capability, and it’s exciting to imagine what others might emerge as we continue to scale up robotic foundation models.

\section*{Acknowledgements}

We thank our robot operators for data collection, evaluations, logistics, and video recording, and our technicians for robot maintenance and repair. We'd like to thank Hunter Hancock for extensive support on the blog post. We also thank Brian Ichter, Karol Hausman, and the entire Physical Intelligence team for feedback throughout the project.

\bibliographystyle{plainnat}
\bibliography{references}

\clearpage
\twocolumn[{
  \centering
  \includegraphics[width=0.9\textwidth]{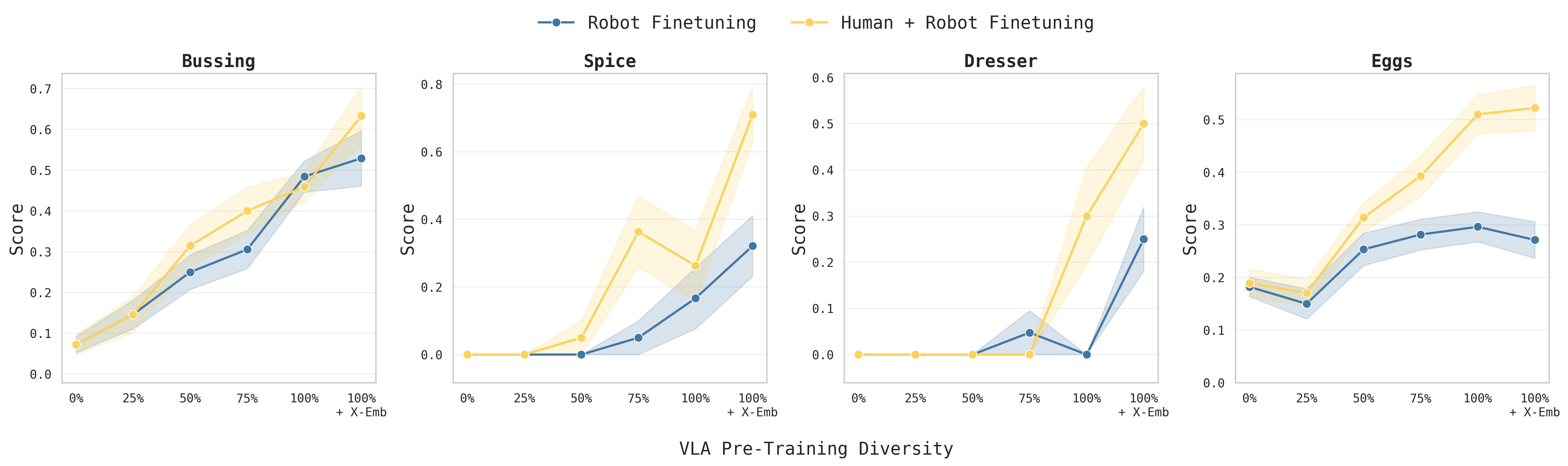}
  \captionof{figure}{\textbf{per task scaling graphs:} Across all tasks we see a clear upward trend in the efficacy of finetuning with human data as pretrained diversity increases.  In \texttt{Eggs} we observe that despite pretraining diversity increasing, the model fails to generalize to this OOD task (blue line), but improves significantly with in-distribution human embodiment data (yellow line).}
  \label{fig:scalingAllLines}
  \vspace{1em} %
}]
\appendix

\subsection{Transfer Benchmark Details}
\label{sec:transfer_benchmark_details}

Our human to robot transfer benchmark consists of 4 tasks, each testing a concept shown only in human data.

\texttt{Bussing:} The robot begins with nine kitchen items on a table, and must place each item in either the trash can or the bussing bin.  The robot earns one point for each item correctly bussed, and we normalize score between [0, 1].

\texttt{Spice:} A kitchen island is set with three randomly placed spice bottles and a spice rack.  The robot earns a point if it successfully places all spice bottles on the rack.  This kitchen is unseen in robot data, but covered in human data.

\texttt{Dresser:} A bedroom dresser is set with a necklace and hair clip.  The robot earns a point by correctly putting the necklace in the jewelry box and the hair clip in the accessories organizer.  This bedroom is unseen in robot data, but covered in human data.

\texttt{Sort Eggs:} The table is set with two egg cartons (half dozen each) and a bowl containing six white and six brown eggs.  The robot scores one point for each white egg it places in the left carton and brown egg in the right carton.  An additional point is awarded for closing each egg carton shut at the end.  Scores are normalized [0, 1].  While the robot data covers picking eggs, the concept of sorting eggs is only covered in human data.

For each experiment, we perform between 20 and 40 evaluations.  All error bars visualize 1 standard error.

\subsection{Results Analysis}
In the context of our experiments, performance improves as a function of two parameters, pretrained model diversity and human data finetuning.  To help disentangle these factors, we can compare the robot-only and human+robot performance scaling curves.

First we can consider the zero shot generalization of our robot only model (blue robot-only curve). For tasks like \texttt{Bussing} and \texttt{Spice}, the zero shot generalization to these new tasks increases steadily with increased pretrained diversity.  However, for \texttt{Dresser} zero shot generalization only emerges in the in our strongest pretrained model.  And for \texttt{Eggs}, performance quickly plateaus even with improved pretraining.  This tells us that increasing pretrained diversity generally improves zero shot generalization across tasks, but it can be nonlinear.

Interestingly, across these tasks there are points in which the zero shot generalization and transferability from human data (yellow Human + Robot curve) are not necessarily correlated.  For instance, with \texttt{Spice} 50\%$\rightarrow$75\% we see a modest increase in zero shot generalization, but a large increase in transferability.  Similarly for Dresser 75\%$\rightarrow$100\% or Eggs 50\%$\rightarrow$100\%+X-emb we see no improvement in zero shot generalization, but a large increase in transfer from human data.  In other words, there are cases where increased pretraining diversity doesn't improve zero shot generalization, but it does improve human to robot transfer.

\subsection{TSNE Visualization}
\label{sec:tsne_details}

To visualize how the model represents the human and robot data internally, we visualize the output embeddings of the VLA from human and robot data with a TSNE in Figure~\ref{fig:tsne}. Specifically, we pass observations from human and robot data through the co-finetuned VLA as input. We then mean-pool the first 200 output embeddings of the VLA (corresponding roughly to ``task'') and visualize them with the TSNE, capturing how well the VLA aligns different observations from humans and robots to the same task. We observe this alignment visibly improves with more diverse pre-training, even after all models have been finetuned on the same data.

\end{document}